\documentclass[10pt]{article}

\usepackage{amsmath,amsthm}
\usepackage{microtype}
\usepackage{charter}
\usepackage{algorithm,algpseudocode}

\def\x{\boldsymbol{x}}
\def\y{\boldsymbol{y}}
\def\u{\boldsymbol{u}}

\newtheorem{theorem}{Theorem}[section]
\newtheorem{proposition}[theorem]{Proposition}

\begin{document}

\title{Constant-time filtering using  shiftable kernels}
\author{Kunal~Narayan~Chaudhury\thanks{kchaudhu@math.princeton.edu}}
\date{}

\maketitle

\begin{abstract}

It was recently demonstrated in  \cite{Chaudhury2011} that the non-linear bilateral filter \cite{Tomasi} can be efficiently implemented using a constant-time or $O(1)$ algorithm. At the heart of this algorithm
was the idea of approximating the Gaussian range kernel of the bilateral filter using trigonometric functions. In this letter, we explain how the idea in \cite{Chaudhury2011} can be extended to few
other linear and non-linear filters \cite{Tomasi,Yaroslavsky,Baudes}. While some of these filters have received a lot of attention in recent years, they are known to be computationally intensive. 
To extend the idea in \cite{Chaudhury2011}, we identify a central property of trigonometric functions, called \textit{shiftability}, that allows us to exploit the redundancy inherent in the filtering
operations. In particular, using shiftable kernels, we show how certain complex filtering can be reduced to simply that of computing the \textit{moving sum} of a stack of images.
Each image in the stack is obtained through an elementary pointwise transform of the input image. This has a two-fold advantage. First, we can use fast recursive algorithms for computing the moving 
sum \cite{Viola2001,Crow}, and, secondly, we can use parallel computation to further speed up the computation. We also show how shiftable kernels can also be used to approximate the (non-shiftable) 
Gaussian kernel that is ubiquitously used in image filtering.

\end{abstract}


\section{Introduction}

The function $\cos(x)$ has the remarkable property that, for any translation $\tau$, $\cos(x-\tau)=\cos(\tau)\cos(x)+\sin(\tau)\sin(x)$. That is, we can express 
the translate of a sinusoid as the linear combination of two fixed sinusoids. More generally, this holds true for  any 
function of the form $\varphi(x)=c_1 \exp(\alpha_1 x)+\cdots+c_N \exp(\alpha_N x)$. 
This follows from the addition-multiplication property of the exponential. As special cases, we have the pure exponentials when $\alpha_i$ is real, and the 
trigonometric functions when $\alpha_i$ is imaginary.

The translation of not all functions can be written in this way, e.g., consider the functions $\exp(-x^2)$ and $(1+x^2)^{-1}$. The other class of functions that have this
property are the polynomials, $\varphi(x)=c_0+c_1 x+\cdots+c_N x^N$. Functions with this property can be realized in higher dimensions using higher-dimensional polynomials and 
exponentials, or simply by taking the tensor product of the one-dimensional functions.

More generally, we say that a function $\varphi(\x)$ is \textit{shiftable} in $\mathbf{R}^d$ if there exists a fixed (finite) collection of 
functions $\varphi_1(\x),\ldots,\varphi_N(\x)$ such that, for every translation $\boldsymbol{\tau}$ in $\mathbf{R}^d$, we can write
\begin{equation}
\label{def}
\varphi(\x-\boldsymbol{\tau}) = c_1(\boldsymbol{\tau}) \varphi_1(\x)+\cdots+c_N(\boldsymbol{\tau}) \varphi_N(\x).
\end{equation}
We call the fixed functions $\varphi_1(\x),\ldots,\varphi_N(\x)$ the \textit{basis functions},  $c_1(\boldsymbol{\tau}),\ldots,c_N(\boldsymbol{\tau})$ the \textit{interpolating coefficients}, and $N$  
the \textit{order} of shiftability. Note that the coefficients depend on $\boldsymbol{\tau}$, and are responsible for capturing the translation action.

Shiftable functions, and more generally, steerable functions \cite{Simoncelli1992}, have a long history in signal and image processing. Over the last few decades, researchers have found applications 
of these special functions in various domains such as image analysis \cite{Freeman1991,Perona1992} and motion estimation \cite{Adelson1985}, and pattern recognition \cite{Watson1985},
to name a few. In several of these applications, the role of steerability and its formal connection with the theory of Lie groups was only recognized much later. We refer the readers to the exposition 
of Teo \cite{Teothesis} for an excellent account on the theory and practice of steerable functions.

In a recent paper \cite{Chaudhury2011}, we showed how specialized trigonometric kernels could be used for fast bilateral filtering. This work was inspired by the work of Porikli \cite{Porikli2008}, who had earlier shown how 
polynomials could be used for the same purpose. We now realize that it is the shiftability of the kernel that is at the heart of the matter, and that this can be 
applied to other forms of filtering and using more general kernels. We will provide a general theory of this in Section \ref{fast_algorithms}, where we also propose some algorithms that have a constant-time complexity per pixel, independent 
of the size of the filtering kernels. To the best of our knowledge, such algorithms have not been reported in the community in its full generality. The problem 
of designing shiftable kernels is addressed in Section \ref{design}. Here we also propose a scheme for approximating the ubiquitous Gaussian kernel using shiftable functions. Finally, in Section \ref{discussion}, 
we present some thoughts on how shiftability could be used for reducing the complexity of the non-local means filter \cite{Baudes}.

\section{Filtering using moving sums}
\label{fast_algorithms}

We now show how certain constant-time algorithms for image filtering can be obtained using shiftable kernels. The idea is that by using shiftability, we can decompose the local kernels (obtained using translations) in 
terms of ``global'' functions -- the basis functions. The local information gets encoded in the interpolating coefficients. 
This allows us to explicitly take advantage of the redundancy in the filtering operation.

To begin with, we consider the simplest neighborhood filter, namely the spatial filter. This is given by
\begin{equation}
\label{filter1}
\overline{f}(\x)=\frac{1}{\eta} \int_{\Omega}  \! \! \varphi(\y)f(\x-\y) \ d\y 
\end{equation}
where $\eta=\int_{\Omega} \! \varphi(\y) \ d\y$. Here $\varphi(\x)$ is the kernel, and $\Omega$ is the neighborhood over which the integral (sum) is taken. Note that, henceforth, we will use the term \textit{kernel} to specifically mean that 
the function is symmetric, non-negative, and unimodal over its support (peak at the origin). It is not immediately clear that one can construct such kernels using shiftable functions. 
We will address this problem in the sequel.

For the moment, suppose that the kernel $\varphi(\x)$ is shiftable, so that \eqref{def} holds. We use this, along with symmetry, to center the kernel at $\x$. In particular, we write
$\varphi(\y)=\varphi(\x-\y-\x)=\sum_{n=1}^N c_n(\x) \varphi_n(\x-\y)$. Using linearity, we have 
\begin{equation*}
\int_{\Omega} \! \!  \varphi(\y)f(\x-\y) \ d\y =\sum_{n=1}^N c_n(\x)  \int_{\Omega}  \varphi_n(\x-\y) f(\x-\y) \ d\y.
\end{equation*}
Similarly, $\eta=\sum_{n=1}^N c_n(\x)  \int_{\Omega}  \varphi_n(\x-\y) \ d\y$.

Now consider the case when $\Omega$ is a square, $\Omega=[-T,T]^2$. Then the above integrals are of the form
\begin{equation*}
\int_{[-T,T]^2}  \! \! F(\x-\y) \ d\y.
\end{equation*}
This is easily recognized as the \textit{moving sum} of $F(\cdot)$ evaluated at $\x$. We will use the 
notation $\mathtt{Sum}(F,\x,T)$ to denote this integral (this is simply the ``moving-average'', but without the normalization). As is well-known, this can be efficiently computed using recursion \cite{Viola2001,Crow}.

The main idea here is that, by using shiftable kernels, we can express \eqref{filter1} using moving sums and these in turn can be computed efficiently. In particular, note that the number of computations
required for the moving sum is independent of $T$, that is, the size of the neighborhood $\Omega$. These are referred to as the constant-time or $O(1)$ algorithms in the image processing community.
The main steps of our  algorithm are summarized in Algorithm \ref{algo_filter1}.

\begin{algorithm}[!htb]
\caption{Constant-time spatial filtering}
\label{algo_filter1}
\begin{algorithmic}
     \State \textbf{Input}: $f(\x)$, $\varphi(\x)$ as in \eqref{def}, and $T$. 
     \State 1. For $1 \leq n \leq N$, use recursion to compute  $\mathtt{Sum}(f \varphi_n,\x,T)$ and $\mathtt{Sum}(\varphi_n,\x,T)$.
     \State 2. Set $\overline{f}(\x)$ as the ratio of $\sum_{n=1}^N c_n(\x) \mathtt{Sum}(f\varphi_n,\x,T)$ and $\sum_{n=1}^N c_n(\x) \mathtt{Sum}(\varphi_n,\x,T)$.
     \State \textbf{Return}: Filtered image $\overline{f}(\x)$.
\end{algorithmic}
\end{algorithm}

The above idea can also be extended to non-linear filters such as the edge-preserving bilateral filter \cite{Tomasi,Yaroslavsky,Brady}. The bilateral filtering of an image $f(\x)$ is given by the formula
\begin{equation}
\label{filter2}
\tilde{f}(\x)=\frac{1}{\eta(\x)} \int_{\Omega} \!\! \varphi(\y) \ \phi(f(\x-\y)-f(\x)) \ f(\x-\y) \ d\y
\end{equation}
where $\eta(\x)=\int_{\Omega}  \varphi(\y) \ \phi(f(\x-\y)-f(\x))  \ d\y$. 

In this formula, the bivariate function $\varphi(\x)$ is called the \textit{spatial kernel}, and the one-dimensional function $\phi(t)$ is called the \textit{range kernel}. Suppose that both these kernels are 
shiftable.
In particular, let $\varphi(\x-\boldsymbol{\tau}) = \sum_{m=1}^{M} c_m(\boldsymbol{\tau}) \varphi_m(\x)$, and $\phi(t-\tau) = \sum_{n=1}^{N} d_n(\tau) \phi_n(t)$.

Plugging these into \eqref{filter2} and using linearity, we can write the numerator as
\begin{equation*}
\sum_{m,n} c_m(\x) d_n(f(\x)) \!\! \int_{\Omega} \varphi_m(\y) \phi_n(f(\x-\y)) f(\x-\y) \ d\y. 
\end{equation*}
Similarly,
\begin{equation*}
\eta(\x)=\sum_{m,n} c_m(\x) d_n(f(\x)) \int_{\Omega} \varphi_m(\x-\y) \phi_n(f(\x-\y)) \ d\y.
\end{equation*}

\begin{algorithm}[!htb]
\caption{Constant-time bilateral filtering}
\label{algo_filter2}
\begin{algorithmic}
     \State \textbf{Input}: $f(\x)$, $\varphi(\x)$ and $\phi(s)$ as in \eqref{filter2}, and $T$. 
     \State 1. For $1 \leq m \leq M$ and $1 \leq n \leq N$, set $a_{m,n}(\x)=c_m(\x) d_n(f(\x)), g_{m,n}(\x)=\varphi_m(\x) \phi_n(f(\x)) f(\x),$ and $h_{m,n}(\x)=\varphi_m(\x) \phi_n(f(\x))$.
     \State 2. Use recursion to compute $\mathtt{Sum}(g_{m,n},\x,T)$ and $\mathtt{Sum}(h_{m,n},\x,T)$.
     \State 3. Set $\tilde{f}(\x)$ as the ratio of $\sum_{m,n} a_{m,n}(\x)\mathtt{Sum}(g_{m,n},\x,T)$ and $\sum_{m,n} a_{m,n}(\x)\mathtt{Sum}(h_{m,n},\x,T)$.
        \State \textbf{Return}: Filtered image $\tilde{f}(\x)$.
\end{algorithmic}
\end{algorithm}	

As before, we again recognize the moving sums when $\Omega=[-T,T]^2$. This gives us a new constant-time algorithm for bilateral filtering, which is summarized in Algorithm \ref{algo_filter2}.
We note that this algorithm is an extension of the one in \cite{Chaudhury2011}, where we used shiftable kernels only for the range filter.

The main computational advantage of Algorithms \ref{algo_filter1} and \ref{algo_filter2} is that the number of computations per pixel does not depend on the size of the spatial or the range kernel. It only depends on the 
order of shiftability. More precisely, the computation time scales linearly with the number of basis functions.   
For example, in case of the bilateral filter, the computation size is roughly $MN$ times the computations required to perform a single moving sum of the image, plus the overhead time of initializing the basis images and 
recombining the outputs. To get an 
estimate of the running time, we implemented Algorithm \ref{algo_filter2} in \texttt{MATLAB}
on an Intel machine with a quad-core 2.83 GHz processor. We found that a single recursive implementation of the moving sum requires roughly $10$ milliseconds on a $512 \times 512$ image when $T=5$. Considering an instance where 
$M\times N=3^2 \times 5$, the total computation time was roughly $2.5$ seconds, $1$ seconds to initialize the basis images and combine the results, and $1.5$ seconds for the moving sums. This indeed looks promising since
we can definitely bring down the time using a \texttt{\texttt{JAVA}} or \texttt{C} compiler. Moreover, we can also use multithreading (parallel computation) to further accelerate the implementation.

\section{Shiftable kernels}
\label{design}

 We now address the problem of designing kernels that are shiftable. The theory of Lie groups can be used to study the class of shiftable functions, e.g., see discussions in \cite{Teothesis,Teoa1998}.
 It is well-known that the polynomials and the exponentials are essentially the only shiftable functions. 

\begin{theorem}[The class of shiftable functions, \cite{Teothesis}]
The only smooth functions that are shiftable are the polynomials and the exponentials, and their sum and product.
\end{theorem}

We recall that a kernel is a smooth function that is symmetric, non-negative, and unimodal. A priori, it is not at all obvious that there exists a kernel that is a polynomial or exponential.
Indeed, the real exponentials cannot form valid kernels since they are neither symmetric nor unimodal. 
On the other hand, while there are plenty of polynomials and trigonometric functions that are both symmetric and non-negative, it is impossible to find a one that 
is unimodal over the entire real line. This is simply because the polynomials blow up at infinity, while  the trigonometric functions are oscillatory.

\begin{proposition}[Conflicting properties]
There is no kernel that is shiftable on the entire real line.
\end{proposition}
 
 Nevertheless, unimodality can be achieved at least on a bounded interval, if not 
the entire real line, using polynomial and trigonometric functions. Note that, in practice, a priori 
bounds on the data are almost always available. For the rest of the paper, and without loss of generality, we fix 
the bounded interval to be $[-T,T]$. 

 We now give two concrete instances of shiftable kernels on $[-T,T]$. The reason for these particular 
choices will be clear in the sequel. For every integer $N=0,1,2,\ldots$, consider the functions
\begin{equation*}
p_N(t)=\left(1-\frac{t^2}{T^2}\right)^N \quad \text{and} \qquad q_N(t)=\left[\cos\left(\frac{\pi t}{2T}\right)\right]^N,
\end{equation*}
where $t$ takes values in $[-T,T]$. It is easily verified that these are valid kernels on $[-T,T]$. The crucial difference between the above kernels is that the
order of shiftability of $p_N(t)$ is $2N+1$, while that of $q_N(t)$ is much lower, namely $N+1$.
We note that it is the kernel $q_N(t)$ that was used in \cite{Chaudhury2011}.

The fact that the sum and product of shiftable kernels is also shiftable can be used to construct kernels in higher dimensions. For example, we can set 
$\varphi(x_1,x_2)=p_N(x_1)p_N(x_2)$, or $\varphi(x_1,x_2)=q_N(x_1)q_N(x_2)$. We call these the \textit{separable kernels}. In higher dimensions, one often requires the kernel to have 
the added property of isotropy. In two dimensions, we can achieve near-isotropy using these separable kernels provided that $N$ is sufficiently large. We will give a precise reason  
 later in the section. However, as shown in a different context in \cite{Chaudhury2010}, it is worth noting that kernels other than the standard separable ones (of same order) can provide better isotropy, 
 particularly for low orders. Indeed, consider the following kernel on the square $[-T,T]^2$:
\begin{equation}
\label{4direction}
\phi(x_1,x_2)= q_1(x_1)q_1\left(\frac{x_1+x_2}{\sqrt{2}}\right)q_1(x_2)q_1\left(\frac{x_1-x_2}{\sqrt{2}}\right).
\end{equation}
The kernel is composed of four cosines distributed uniformly over the circle. It can be verified that $\phi(x_1,x_2)$ is more isotropic 
than the separable kernel of same order, $q_2(x_1)q_2(x_2)$. However, note that the non-negativity constraint is violated on the corners of the square $[-T,T]^2$ in this case. This is unavoidable 
since we are trying to approximate a circle within a square. Nevertheless, this does not create much of a problem since the negative overshoot is well within $2\%$ of the peak value. Following the 
same argument, the polynomial
\begin{equation*}
\phi(x_1,x_2)=p_1(x_1)p_1\left(\frac{x_1+x_2}{\sqrt{2}}\right)p_1(x_2)p_1\left(\frac{x_1-x_2}{\sqrt{2}}\right)
\end{equation*}
tends to be more isotropic on $[-T,T]^2$ than $p_2(x_1)p_2(x_2)$.

\subsection{Approximation of Gaussian kernels}

The most commonly used kernel in image processing is the Gaussian kernel. This kernel, however, is not shiftable. As a result, we cannot directly apply our algorithm for the Gaussian kernel.
One straightforward option is to instead approximate the Gaussian using its Fourier series or its Taylor polynomial, both of which are shiftable. The difficulty with either of these is that
they do not yield valid kernels. That is, it is not guaranteed that the resulting approximation is non-negative or unimodal; see \cite[Fig. 1]{Chaudhury2011}. This is exactly the problem with the polynomial approximation
used for the bilateral filter in \cite{Porikli2008}. As against these, we can instead use the specialized shiftable kernels $p_N(t)$ and $q_N(t)$ to approximate the Gaussian to any arbitrarily precision, based on the following results.

\begin{theorem}[Gaussian approximation] For every $-T \leq t \leq T$, 
\begin{equation}
\label{approx1}
\lim_{N \longrightarrow \infty}  p_N\left(\frac{t}{\sqrt{N}}\right)=\exp\left(-\frac{t^2}{T^2}\right),
\end{equation}
and
\begin{equation}
\label{approx2}
\lim_{N \longrightarrow \infty}  q_N\left(\frac{t}{\sqrt{N}}\right)=\exp\left(-\frac{\pi^2 t^2}{8T^2}\right).
\end{equation}
\end{theorem}
In either case, the convergence takes place quite rapidly. The first of these results is well-known. For a proof of the second result, we refer the 
readers to \cite[Sec. II-D]{Chaudhury2011}. The added utility of these asymptotic formulas is that we can control the variance of these kernels using the variance of the target Gaussian. This is particularly useful because
no simple closed-form expressions for the variance of $p_N(t)$ and $q_N(t)$ are available.
We refer the authors to \cite[Sec. II-E]{Chaudhury2011} for details on how one can exactly control the variance of $q_N(t)$. The same idea applies to $p_N(t)$.

We now discuss how to approximate isotropic Gaussians in two dimensions using shiftable kernels. Doing this using separable kernels is straightforward. For example,
\begin{equation}
\label{separable_conv}
\lim_{N \longrightarrow \infty}  q_N\left(\frac{x_1}{\sqrt{N}}\right)q_N\left(\frac{x_2}{\sqrt{N}}\right)=\exp\left(-\frac{\pi(x_1^2+x^2_2)}{8T^2}\right).
\end{equation}
There is yet another form of convergence which is worth mentioning. Consider the following kernel defined on $[-T,T]^2$:
\begin{equation*}
\phi_N(x_1,x_2)=\prod_{k=1}^N q_1\Big(\sqrt{\frac{6}{N}} (x_1 \cos\theta_k +x_2\sin\theta_k) \Big),
\end{equation*}
where $\theta_k=(k-1)\pi/N$. The kernel $\phi_4(x_1,x_2)$ is simply the (rescaled) kernel $\phi(x_1,x_2)$ in \eqref{4direction}. By slightly adapting the proof of Theorem 2.2 in \cite[Appendix A]{Chaudhury2010}, we can show 
the following.

\begin{theorem}[Approximation of isotropic Gaussian] For every $(x_1,x_2)$ in $[-T,T]^2$,
\begin{equation*}
\lim_{N \longrightarrow \infty} \phi_N(x_1,x_2)=\exp\left(-\frac{\pi^2(x_1^2+x^2_2)}{8T^2}\right).
\end{equation*} 
\end{theorem}
Note that the target Gaussian in this case is the same as in \eqref{separable_conv}. The key difference, however, is that for low orders $N$, e.g. $N=4$, $\phi_N(x_1,x_2)$ looks more isotropic than 
$ q_N(x_1/\sqrt{N})q_N(x_2/\sqrt{N})$. However, as noted earlier, the non-negativity requirement of a kernel is mildly violated by $\phi_N(x_1,x_2)$ at the corners of the square domain.

The significance of the order $N$ is that it allows one to arbitrarily control the accuracy of the Gaussian approximation. As discussed in \cite[Sec. II-E]{Chaudhury2011}, $N$ has to be greater than a threshold $N_0$
for the  approximating kernels in \eqref{approx1} and \eqref{approx2} to be non-negative and unimodal. In particular, if $\sigma^2$ is the variance
of the target Gaussian, then $N_0$ is of the order $O(T^2/\sigma^2)$. Thus, a large $N_0$ is required to approximate a narrow Gaussian on a large interval. For the spatial kernel, the ratio $T^2/\sigma^2$ is usually 
small, since $T$ is small in this case. This is, however, not the the case for the range kernel of the bilateral filter, where $T$ can almost be as large as the dynamic range of the image. The good news is that, by allowing for mild (and controlled) violations of the non-negativity constraint, one can closely approximate the Gaussian using a significantly lower number of terms. This is 
done by discarding the less significant basis functions in \eqref{def}; see \cite{AppendixChaudhury2011} for detailed discussion and results. The same idea can also be applied to the polynomials.

\section{Discussion}
\label{discussion}

We have shown how, using shiftable kernels, we can express two popular forms of image filters (and possibly many more) in terms of simple moving sums.
We note that we can speed up the implementation of Algorithm \ref{algo_filter1} and \ref{algo_filter2} using parallel computation or multithreading.
In future work, we plan to implement these algorithms in \texttt{C} or \texttt{JAVA}, and make extensive comparison of the result and execution 
time with the state-of-the-art algorithms.

To conclude, we note that the main idea can also be extended to some other forms of neighborhood filtering, e.g., the ones in \cite{Yaroslavsky,Brady}. What is even more interesting is that we can extend the 
idea for approximating the non-local means filter \cite{Baudes}. The non-local means is a higher order generalization of the bilateral filter, where one works with patches instead of single pixels.
The filtered image $\hat{f}(\x)$ in this case is given by
\begin{equation}
\label{NL}
\hat{f}(\x)= \frac{ \int  f(\x-\y) w(\x,\y) \ d\y}{  \int  w(\x,\y) \ d\y}
\end{equation}
where 
\begin{equation*}
 w(\x,\y)=\exp(-h^{-2} \int g(\u) \big(f(\x+\u)-f(\x-\y+\u)\big)^2 d\u)
\end{equation*}
Here $g(\u)$ is a two-dimensional Gaussian, and the integrals (sums)  are taken over the entire image domain. In practice, though, the sum is performed locally \cite{Baudes}.

It is possible to express \eqref{NL} in terms of moving sums, using shiftable approximates of the Gaussian. First, we approximate the domain of the outer integral by
a sufficiently large square $[-T,T]^2$, and the inner integral by a finite sum over $p$ neighborhood pixels $u_1,\ldots,\u_p$, where, say, $\u_1=0$. In this case, $\hat{f}(\x)$ is
given by
\begin{align}
\label{approx_multivariate}
\frac{1}{\eta(\x)} \int_{[-T,T]^2} \!\! & f(\x-\y) \varphi(f(\x+\u_1)-f(\x-\y+\u_1),\ldots \nonumber \\
&\ldots,f(\x+\u_p)-f(\x-\y+\u_p))  d\y
\end{align}
where $\varphi(t_1,\ldots,t_p)=\exp(-h^{-2} \sum_{k=1}^p g(\u_k) t_k^2)$, and where $\eta(\x)$ is given by
\begin{align}
\label{inetgrand}
\int_{[-T,T]^2}   \!\! \! & \varphi(f(\x+\u_1)-f(\x-\y+\u_1),\ldots \nonumber \\
&\ldots,f(\x+\u_p)-f(\x-\y+\u_p))  d\y.
\end{align}
Note that $\varphi(t_1,\ldots,t_p)$ is an anisotropic Gaussian in $p$ variables with covariance $\mathtt{diag}(h^2/2g(\u_1),\ldots,h^2/2g(\u_p))$. Now, using a shiftable 
approximation (we continue using the same symbol) of $\varphi(t_1,\ldots,t_p)$ as in \eqref{def}, we can write the integrand in \eqref{inetgrand} as 
$\sum_{n=1}^N c_n (f(\x+\u_1),\ldots,f(\x+\u_p))  \varphi_n(f(\x-\y+\u_1),\ldots,f(\x-\y+\u_p))$. 

We can then write the numerator in \eqref{approx_multivariate} as $\sum_n c_n(f(\x+\u_1),\ldots,f(\x+\u_p)) \mathtt{Sum}(G_n,\x,T)$, where 
we set $G_n(\x)= f(\x)\varphi_n(f(\x+\u_1),\ldots,f(\x+\u_p))$. Similarly, letting $H_n(\x)= \varphi_n(f(\x+\u_1),\ldots,f(\x+\u_p))$, we have $
\eta(\x)=\sum_n c_n(f(\x+\u_1),\ldots,f(\x+\u_p)) \mathtt{Sum}(H_n,\x,T)$.

The catch here is that it get a good approximation of non-local means we need to make both $T$ and $p$ large. While there is no problem in making $T$ large 
(the cost of the moving sum is independent of $T$), it is rather difficult to make $p$ large. For example, with a separable approximation of  $\varphi(t_1,\ldots,t_p)$,  
the overall order $N$ would scale as $n^p$, where $n$ is the approximation order of the Gaussian along each dimension. This limits the scheme to coarse approximations, and to small neighborhoods. 
To make this practical, we need a a polynomial or trigonometric approximation of $\varphi(t_1,\ldots,t_p)$ whose order grows slowly with $p$. It would indeed be interesting to see 
if such approximations exist at all.

\section{Acknowledgment}

The author would like to thank Michael Unser for reading the manuscript and for his useful comments. The author was supported by a fellowship from the Swiss National Science Foundation under grant PBELP2-$135867$.

\bibliographystyle{plain}
\bibliography{bibliography}

\end{document}